\documentclass[final]{cvpr}

\usepackage{times}
\usepackage{epsfig}
\usepackage{graphicx}
\usepackage{amsmath}
\usepackage{amssymb}
\usepackage{mathrsfs}
\usepackage{booktabs}
\usepackage{multirow}
\usepackage[ruled,vlined]{algorithm2e}
\DeclareMathAlphabet{\mathscrbf}{OMS}{mdugm}{b}{n}

\usepackage[pagebackref=true,breaklinks=true,colorlinks,bookmarks=false]{hyperref}

\def\cvprPaperID{2940} 
\def\confYear{CVPR 2021}
\begin{document}

\title{FlowStep3D: Model Unrolling for Self-Supervised Scene Flow Estimation} 

\author{
\hspace{3mm}Yair Kittenplon\textsuperscript{1} \qquad Yonina C. Eldar\textsuperscript{2} \qquad Dan Raviv\textsuperscript{1}
\\
\textsuperscript{1}Tel Aviv University \hspace{5mm} \textsuperscript{2}Weizmann Institute of Science

\\ {\tt\small \hspace{5mm} \{yairk@mai1, darav@tauex\}.tau.ac.il \hspace{4 mm} yonina.eldar@weizmann.ac.il \hspace{17 mm}}
}

\maketitle

\begin{abstract}
Estimating the 3D motion of points in a scene, known as scene flow, is a core problem in computer vision.
Traditional learning-based methods designed to learn end-to-end 3D flow often suffer from poor generalization. Here we present a recurrent architecture that learns a single step of an unrolled iterative alignment procedure for refining scene flow predictions.
Inspired by classical algorithms, we demonstrate iterative convergence toward the solution using strong regularization.
The proposed method can handle sizeable temporal deformations and suggests a slimmer architecture than competitive all-to-all correlation approaches. Trained on FlyingThings3D synthetic data only, our network successfully generalizes to real scans, outperforming all existing methods by a large margin on the KITTI self-supervised benchmark.\footnote {\url{https://github.com/yairkit/flowstep3d}}

\end{abstract}

\section{Introduction}
Understanding motion is fundamental to many applications in a variety of fields, such as human-computer interaction, robotics, and autonomous driving.
The information absorbed within a temporal window is not only a collection of images or a representation of an outcome, but also a description of a process.

Decades ago, computer vision tackled the task of motion estimation, searching for a flow between two images \cite{broxmalik, flownet, hornshunk, lk, deepflow}.
\begin{figure}[t]
\begin{center}
\includegraphics[width=0.98\linewidth]{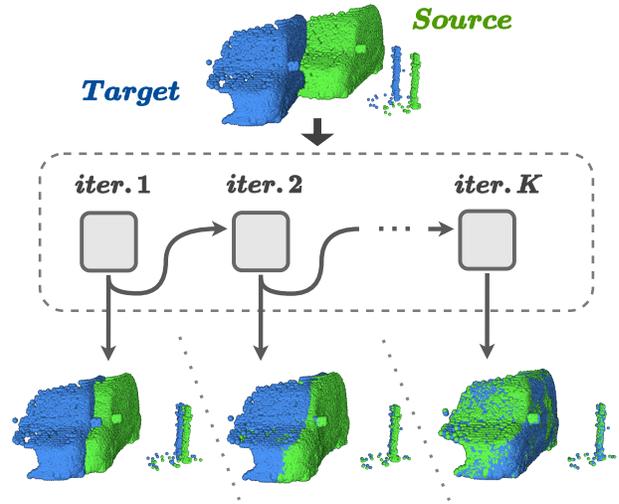}
\end{center}
   \caption{\textbf{Model unrolling.} Top: source (green) and target (blue) input point clouds sampled from the KITTI \cite{kitti2018} scene-flow dataset. Middle: illustration of our FlowStep3D architecture unrolled for K iterations. Bottom: source warped over the predicted flow at each iteration toward the static target at inference time.}
\label{fig:teaser}
\end{figure} 
One significant leap forward in understanding the motion of a scene, defined as scene flow, is the presence of 3D geometry. It liberates us from considering color as the main correspondence feature and allows examining the structure itself to understand the motion. Axiomatic concepts of rigidity \cite{rigidicp, 2016rigid} provided fast and accurate results, but once piece-wise movements \cite{picewiseicp, piecewisesf, 2013piecewise} or non-rigidity \cite{nicp, nrigidreg2, nrigidreg} was allowed, scene flow estimation problem became ill-posed and unfortunately hard to solve.

The rise of artificial intelligence \cite{lecun1989} gives hope that solving the 3D flow estimation problem is possible using a network architecture. Indeed, in the last few years, an improvement in different learning-based methods \cite{HPLFlowNet,flownet3d, FLOT, Ego, pointPWC} has been seen, outperforming those that relied on optimization. More importantly, these learned models are fast and robust.

Scene flow estimation is an integral component in the autonomous driving industry, where LiDAR data is used for the perception of the environment. However, LiDAR sensors suffer from sparseness, directly affecting deep-learning flow algorithms that require knowledge of the objects' spatio-temporal neighborhood. In other words, once the structures do not heavily overlap, the process fails.
In an attempt to solve this limitation, we recently have seen all-to-all mechanisms both for images \cite{raft} and geometry \cite{FLOT}. However, these methods consume large amounts of memory and tend to produce outliers, as now nearby points can be aligned with inconsistent temporal positions.

In this work, we focus on the scene flow problem, where large deviations between the scenes can occur. A small set of points is used to guide the alignment in an all-to-all approach, and a recurrent refinement block is then unrolled to learn movement differentiators.
We train our network to predict a single step at a time and converge iteratively toward the end flow solution, as illustrated in Fig.\ref{fig:teaser}.
Although unrolled for K iterations during training, our network can be used for inference with a larger number of iterations to handle more significant and complicated deformations.

Trained on synthetic data only, our method improves the state-of-the-art results on the self-supervised KITTI benchmark by a considerable margin. Our architecture is further tested in a fully-supervised framework
and achieves slightly better results compared to prior art while benefiting from memory efficiency.

The key contributions of this work are as follows:
\begin{itemize}
    \item We present the first recurrent architecture for non-rigid scene flow.
    \item We provide a slim memory all-to-all correlation pipeline by merging low-resolution correlation with an unrolling iterative refinement process.
    \item Our proposed network achieves large improvements over existing self-supervised methods on both FlyingThings3D and KITTI benchmarks.

\end{itemize}
\section{Related Work}
\noindent\textbf{Scene Flow Estimation on Point Clouds.} 
\space Scene flow estimation was first introduced in \cite{Vedula}, who suggested to compute a 3D scene flow from 2D optical flow using a linear algorithm. 
Later approaches used stereo sequences \cite{Variational}, RGB-D \cite{RGBD-sf}, and LiDAR \cite{2016rigid}.
With the rise of new methods for deep learning on point clouds \cite{pointnet, pointnet++, splatnet, pointconv} and the increasing popularity of range data at the autonomous driving domain, more recent approaches suggest learning the 3D scene flow directly from the raw data spatial positions.

Liu \etal \cite{flownet3d} were the first to introduce a correlation layer that aggregates features of different point clouds based on PointNet\cite{pointnet}. However, the correlation layer was applied at a particular scale, only capturing the correlation of that specific level of features between the point clouds and a fixed neighborhood radius, allowing a small deformation between the point clouds. Gu \etal\cite{HPLFlowNet} tackled those limitations by introducing multi-resolution correlation layers and suggested using a Bilateral Convolutional layer \cite{bcl1, bcl2, splatnet}. Inspired by classical pyramid approaches, Wu \etal \cite{pointPWC} further improved multi resolution flows by applying it in a coarse to fine manner and showed superior results.
However, multi-resolution methods require many learnable parameters and are limited to deformations smaller than the correlation neighborhood.
In \cite{Ego}, the authors suggested splitting the movement into rigid ego-motion and non-rigid refinement components, relying on the same architecture as \cite{HPLFlowNet}.
A different approach focusing on all-to-all correlation suggested by \cite{FLOT}, using optimal transport tools to estimate the scene flow, showed excellent results.
However, an all-to-all correlation matrix for a large scale point cloud is inefficient. 

We adopt the all-to-all correlation concept, but unlike \cite{FLOT}, we suggest to use it efficiently in a much deeper, lower resolution space.

\begin{figure}[t]
\begin{center}
\includegraphics[width=0.95\linewidth]{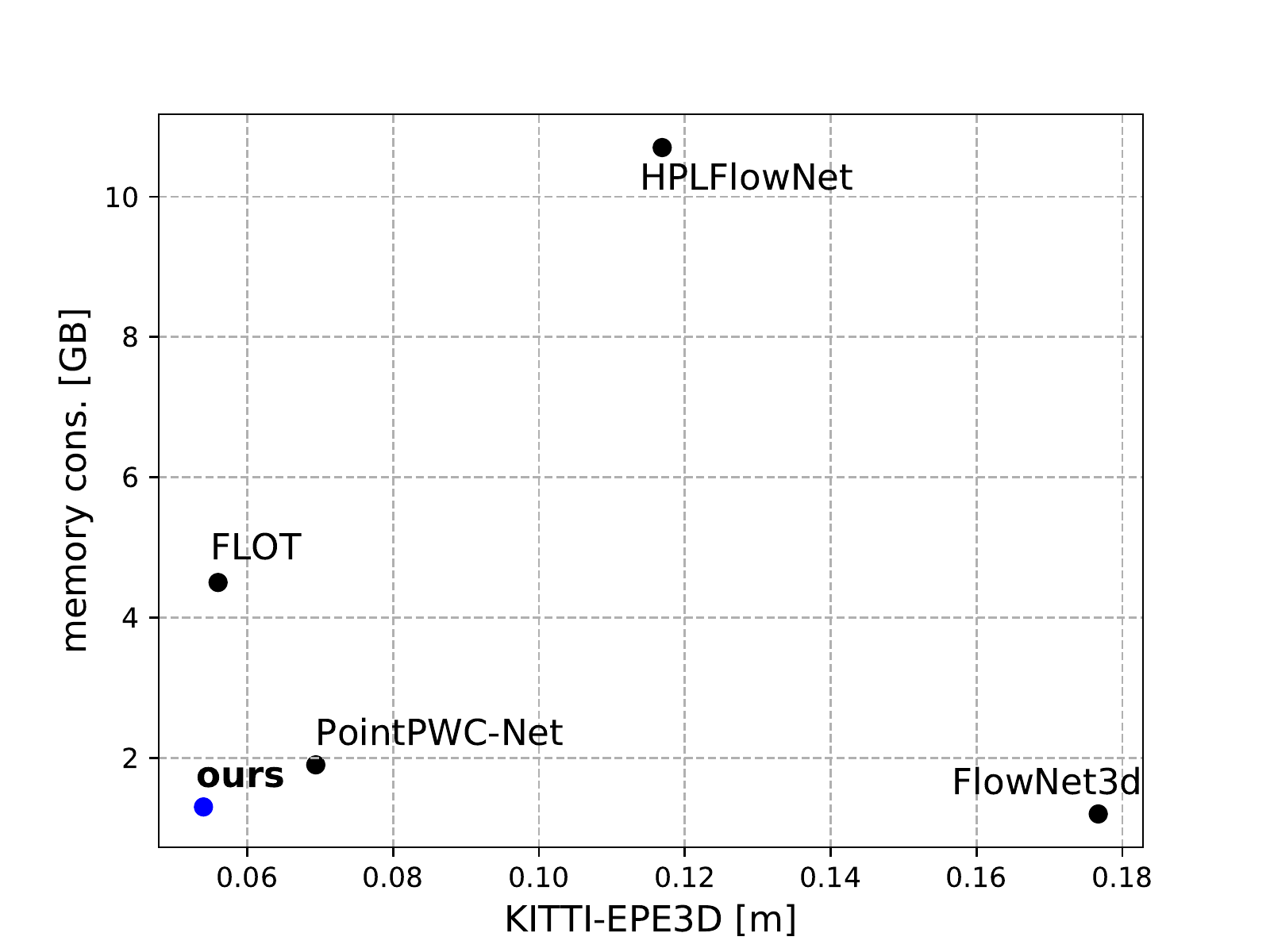}
\end{center}
   \caption{\textbf{Memory consumption and accuracy trade-off.} Average end-point-error (Sec.\ref{sec:exp}) of the leading fully-supervised methods on the KITTI test vs. memory consumption on inference with 8192 points per scene.}
\label{fig:diagram}
\end{figure}

\begin{figure*}
\begin{center}
\includegraphics[width=0.95\linewidth]{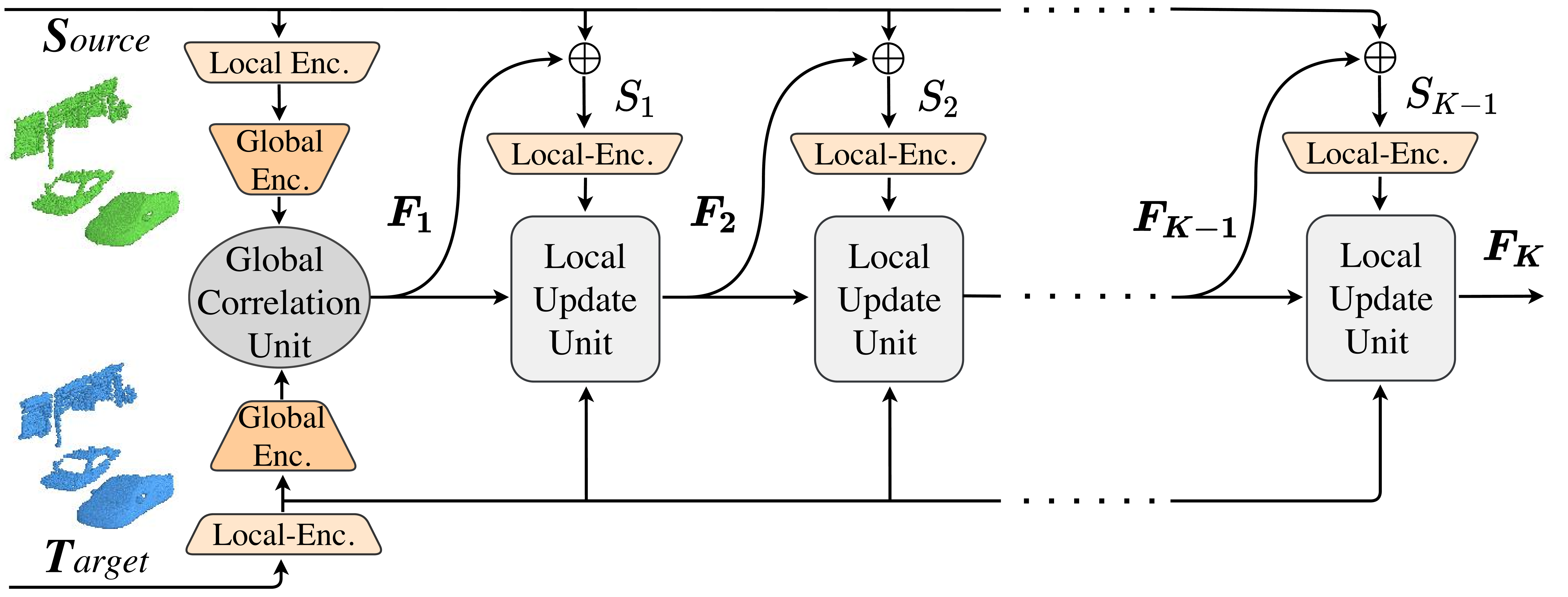}
\end{center}
\vspace{-1.5mm} 
   \caption{\textbf{FlowStep3D high-level overview.} At the first iteration, the global correlation unit (Fig.~\ref{fig:global_unit}) produces the initial flow $F_1$ based on source's and target's global features obtained by deep encoding. In each iteration, the source point cloud is warped toward the target by adding the predicted flow from the previous iteration $F_{k-1}$. It is then locally encoded and fed into a local update unit (Fig.~\ref{fig:local_update_unit}) to refine the flow estimation. The weights of the local update unit and of the encoders are shared across all their appearances.}
\label{fig:system_overview}
\end{figure*}

\noindent\textbf{Self-supervised Learning.} 
\space Learning to estimate scene flow in a self-supervised manner is an active field of research. Mittal \etal \cite{JustGW} showed that cycle-consistency and nearest-neighbor losses could be used for self-supervision of scene flow learning, using a backbone of FlowNet3D \cite{flownet3d}, pre-trained in a fully supervised manner on synthetic data. Tishchenko  \etal \cite{Ego} combined the same self-supervised losses with a fully supervised loss into a 'Hybrid loss', while \cite{pointPWC} suggested a fully self-supervised process, combining Chamfer, nearest-neighbors, and laplacian losses.

We follow \cite{pointPWC} and choose the Chamfer loss as the data-term loss of our training. Still, inspired by classical non-rigid alignment algorithms, we claim that this data term is not sufficient for a one-shot, end-to-end solution. Hence we suggest an iterative approach for the scene flow estimation and emphasize the need for strong regularization loss term.

\noindent\textbf{Algorithm Unrolling.} 
\space While the vast majority of deep learning approaches propose a purely data-driven, one-shot solution, there is a rising trend of combining iterative algorithms to neural network architectures to take advantage of both learning and prior knowledge.
Recent works showed promising results for signal and image processing tasks \cite{viterbinet, unrol_img_de, learningproxunrol, metzlerunrol, unrol_graph_de, unrol_osnet, unrol_usr} by unrolling either an explicit iterative solution for an energy minimization problem or a model.
A contemporary approach named RAFT \cite{raft} suggested model unrolling for 2D optical flow estimation, performing lookups on a 4D all-to-all correlation volume.

We suggest to unroll a single-step flow estimation model. Inspired by \cite{raft}, we also adopt the idea of using a gated recurrent unit for iterative updates. An essential concept of our method, which is different from \cite{raft}, is the computation of new features for the warped scene at every iteration. It is necessary since all point cloud convolution methods are not rotation invariant, so the features of the source change as it is being rotated toward the target. We consider this process as a critical component to learning differentiators iteratively.

\section{Problem Definition}
Scene flow is the 3D motion field of points in a scene.
For a given two sets of points $S=\{p_i\in {\mathbb R}^3\}_{i=1}^{n_1}$ and $T=\{q_j\in {\mathbb R}^3\}_{j=1}^{n_2}$, sampled from a dynamic scene at two consecutive time frames, we denote by $f_i\in {\mathbb R}^3$ the translational motion vector of a point $p_i\in S$ from the first frame toward its new location in the second frame.
Our goal is to estimate the scene flow $F=\{f_i\}_{i=1}^{n_1}$ that describes the best non-rigid transformation, which aligns $S$ toward $T$.
Due to both the sparsity of the 3D data and possible occlusions, a point $p'_i$ may not be presented in $T$. Therefore, we do not learn the correspondence between $S$ and $T$, but a flow representation for each point $p_i\in S$.

In general, every point $p_i, q_j$, may have additional information such as color or geometric features. The number of points in the source may differ from the number of points in the target, i.e., $n_1$ and $n_2$ are not necessarily equal.
\section{Architecture}
We suggest an iterative system (Fig.~\ref{fig:system_overview}) that predicts a flow sequence $\{\boldsymbol{F}_1,...,\boldsymbol{F}_{K}\}$, where $\boldsymbol{F}_{K} = \boldsymbol{F^*}$  is our final flow estimation.
First, we use a global correlation unit (Sec.~\ref{sec:global}) to guide the alignment in an all-to-all approach.
Next, we unroll a local update unit (Sec.~\ref{sec:local}), to learn movement refinements.
Our local update unit implements a single conceptual iteration of an Iterative-Closest-Point (ICP) algorithm \cite{medioni, rigidicp}, replacing the two phases (a. finding correspondence and b. estimating the best smooth transformation based on that correspondence) by learned components.

The number of iterations K is a hyper-parameter and can be larger during inference than during training to handle more complicated and large deformations, as discussed in Sec.~\ref{sec:ablation}.

\subsection{Local And Global Features Encoding}\label{sec:encoding}
Local features of a point encode the geometric features of its relatively small neighborhood and are useful for local alignment refinements. On the other hand, global features capture high-level information regarding the relative position of the point in the scene, using a larger receptive field and deeper encoding.
A crucial part of our method is the distinction between the local and the global features of a point cloud.

We use the \textit{set\_conv} layer suggested by FlowNet3D \cite{flownet3d} as our convolution mechanism and \textit{furthest\_point\_sampling} method for down-sampling.
Our local encoder $g_\theta: {\mathbb R}^{n\times3} \mapsto {\mathbb R}^{n'\times d_{local}}$ consists of only two \textit{set\_conv} layers, capturing a relatively small receptive field, so that its output encodes an input point clouds shallow features of dimension $d_{local}$, at resolution $n'$.
Local encoding is first applied on both source and target input point clouds to produce $g_\theta(S)$, $g_\theta(T)$, and then applied again at every iteration $k$ on the warped source point cloud $S_k$ producing $g_\theta(S_k)$.

In order to extract global features, the local features descriptors $g_\theta(S)$, $g_\theta(T)$ are injected into an additional encoder $h_\theta:  {\mathbb R}^{n'\times d_{local}} \mapsto {\mathbb R}^{n''\times d_{global}}$, which produces $h_\theta(g_\theta(S))$ and $h_\theta(g_\theta(T))$, a deeper representation of $S$ and $T$ of dimension $d_{global}$ and resolution $n''<<n'$, which we denote as $\tilde{h}_\theta(S)$, $\tilde{h}_\theta(T)$ to ease notations.
Both $g_\theta$ and $h_\theta$ encoders have shared weights across all their appearances.

\subsection{Global Correlation Unit}\label{sec:global}
\begin{figure}[t]
\begin{center}
\includegraphics[width=0.9\linewidth]{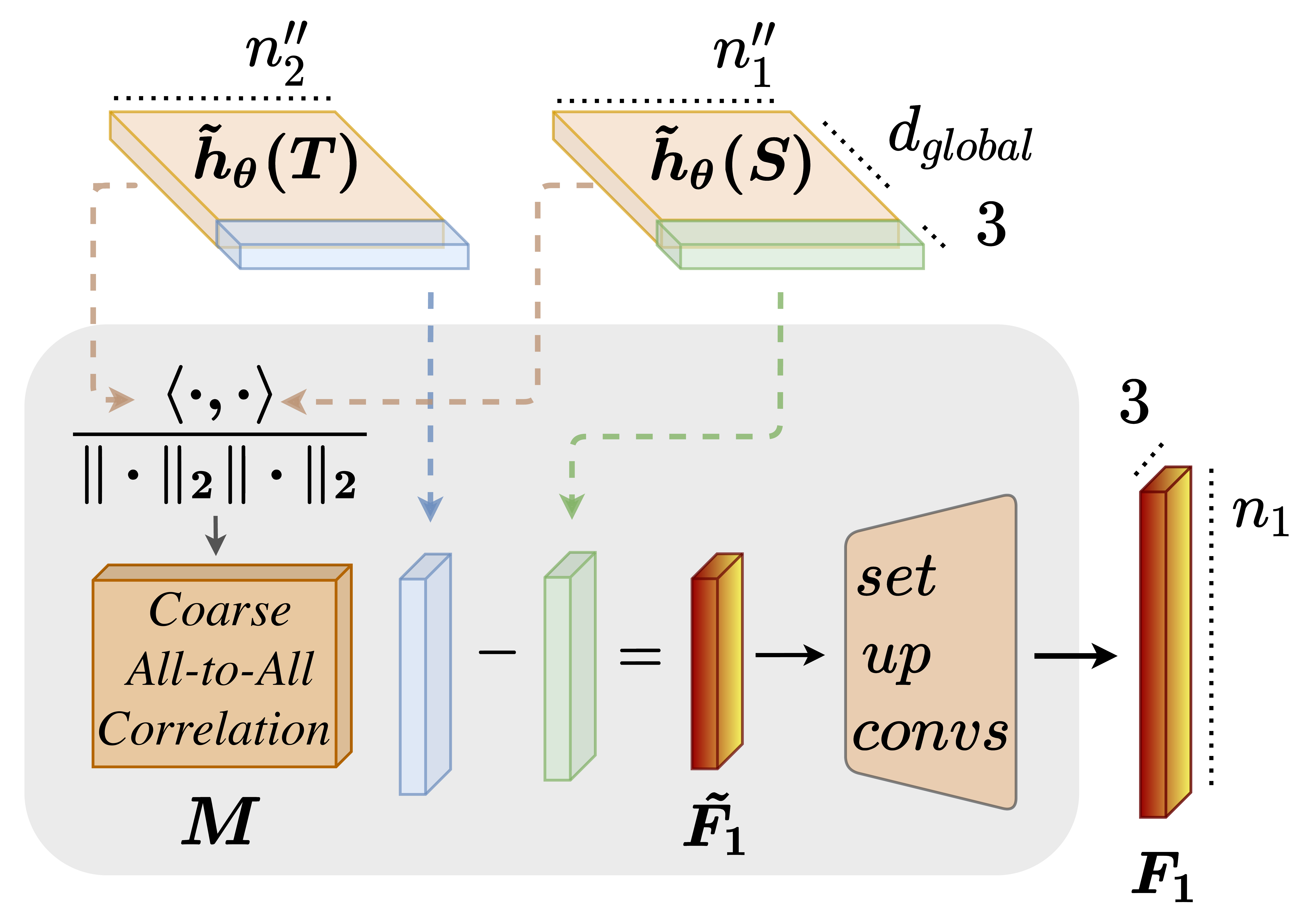}
\end{center}
   \caption{\textbf{Global Correlation Unit.} 
   The inner-product of  $\tilde{h}_\theta(S)$ and $\tilde{h}_\theta(T)$, the global features of the source and target, is used to create coarse all-to-all correlation matrix M. Matrix multiplication followed by a set of \textit{set\_up\_conv} layers are then used to predict the global flow $F_1$, as described in Sec. \hyperref[sec:global]{4.2}.}
\label{fig:global_unit}
\end{figure}

We use a global correlation unit to estimate the initial scene flow $\boldsymbol{F_1}$ based on a deep, coarse all-to-all mechanism, illustrated in Fig. ~\ref{fig:global_unit}.

\noindent\textbf{Coarse All-to-all Correlation Matrix.} \space As the first step of our global correlation unit, we use $\tilde{h}_\theta(S)$, $\tilde{h}_\theta(T)$ to calculate a coarse all-to-all correlation matrix $M \in {\mathbb R}^{n_1''\times n_2''}$.
Inspired by FLOT \cite{FLOT}, we calculate cosine similarity between the features vectors:
\begin{equation}
{sim(i, j)} = \frac{{\tilde{h}_\theta(p)_i^T}{\tilde{h}_\theta(q)}_j}{\|{\tilde{h}_{\theta}(p)_i}\|_2\|{\tilde{h}_{\theta}(q)_j}\|_2} ,
\end{equation}
and then use an exponential function to derive from it a soft correlation matrix:
\begin{equation}
M_{i,j} = \exp\left({\frac{{sim(i, j)} - 1}{\epsilon}}\right).
\end{equation}
Thus, every entry $M_{i,j}$, describes the correlation between ${\tilde{h}_\theta}{(p)_i}$ and ${\tilde{h}_\theta}{(q)_j}$, and the softmax temperature $\epsilon$ is a hyper-parameter, set to $0.03$.

Unlike \cite{FLOT}, we calculate the all-to-all correlation matrix $M$ at a lower dimension, $n_1''\times n_2''<<n_1\times n_2$, thereby significantly reduce the required memory.

\noindent\textbf{Global Flow Estimation.}
\space In order to use the calculated correlation matrix for global flow embedding in the euclidean space, we apply a simple matrix multiplication:
\begin{equation}
\tilde{f_i}= \frac{\sum_{j=1}^{n_2''} M_{i,j}\tilde{q_j}}{\sum_{j=1}^{n_2''} {M_{i,j}}}-\tilde{p_i}.
\end{equation}
Thus, $\tilde{f_i}$ is the average distance between $\tilde{p_i} \in \tilde{S}$  to all points $\{\tilde{q_j}\}_{j=1}^{n_2''} \in\tilde{T}$, weighted by their correlation magnitudes, where $\tilde{S}\in {\mathbb R}^{{n_1''}\times 3}$ and $\tilde{T}\in {\mathbb R}^{{n_2''}\times 3}$ are the coarse versions of $S$ and $T$ coordinates after the the encoder's down-sampling.
The first iteration flow $\boldsymbol{F_1}\in {\mathbb R}^{n_1\times3}$, is regressed out of $\boldsymbol{\tilde{F_1}}=\{\tilde{f_i}\}_{i=1}^{n_1''}$ by a set of \textit{set\_up\_conv} layers \cite{pointnet++}.
\subsection{Local Update Unit}\label{sec:local}
We use an iterative update procedure, starting at the global flow estimation $\boldsymbol{F_1}$, and estimating the rest of the flow sequence $\{\boldsymbol{F}_2,...,\boldsymbol{F}_{K}\}$ based on local information.

\noindent\textbf{Warp and Encode.} 
\space At each iteration $k\in \{2,..,K\}$, we use the estimated flow from the previous iteration for warping the points of the source, i.e $S_{k-1} = S + F_{k-1}$.
Next, using the local encoder $g_\theta$, we extract a new local features descriptor for the warped source  $g_\theta(S_{k-1})$, which we will later use for local correlation calculation (Fig.~\ref{fig:system_overview} top).

\noindent\textbf{Local Correlation.} 
\space To derive the correlation between the local features of the warped source and the target, we adopt the \textit{flow\_embedding} correlation layer proposed by FlowNet3D \cite{flownet3d}.
The proposed correlation layer aggregates feature similarity and spatial relationships of points within a local neighborhood, and therefore is suitable for local refinements.
Specifically, at each iteration $k$, we calculate $\textit{flow\_embedding}(g_\theta(S_{k-1}), g_\theta(T))\in {\mathbb R}^{{n_1'}\times d_{corr}}$, which encodes flow embedding for every point in the warped source $S_{k-1} $ toward the target $T$.

\noindent\textbf{Gated Recurrent Unit (GRU).} 
\space Inspired by RAFT \cite{raft}, we use a gated activation unit based on the design of a GRU cell \cite{gru} as our updating mechanism.
Given previous iteration hidden state $h_{k-1}$, together with current iteration information $x_k$, it produces an updated hidden state $h_k$:
\begin{equation}
z_k = \sigma(set\_conv_z([h_{k-1}, x_k])),
\end{equation}
\begin{equation}
r_k = \sigma(set\_conv_r([h_{k-1}, x_k])),
\end{equation}
\begin{equation}
\tilde{h}_k = tanh(set\_conv_h([r_k\odot h_{k-1}, x_k])),
\end{equation}
\begin{equation}
h_k = (1-z_k)\odot h_{k-1} + z_k\odot \tilde{h}_k,
\end{equation}
where $\odot$ is the Hadamard product, $[\cdot,\cdot]$ is a concatenation and $\sigma(\cdot)$ is the sigmoid activation function.

\begin{figure}[t]
\begin{center}
\includegraphics[width=0.98\linewidth]{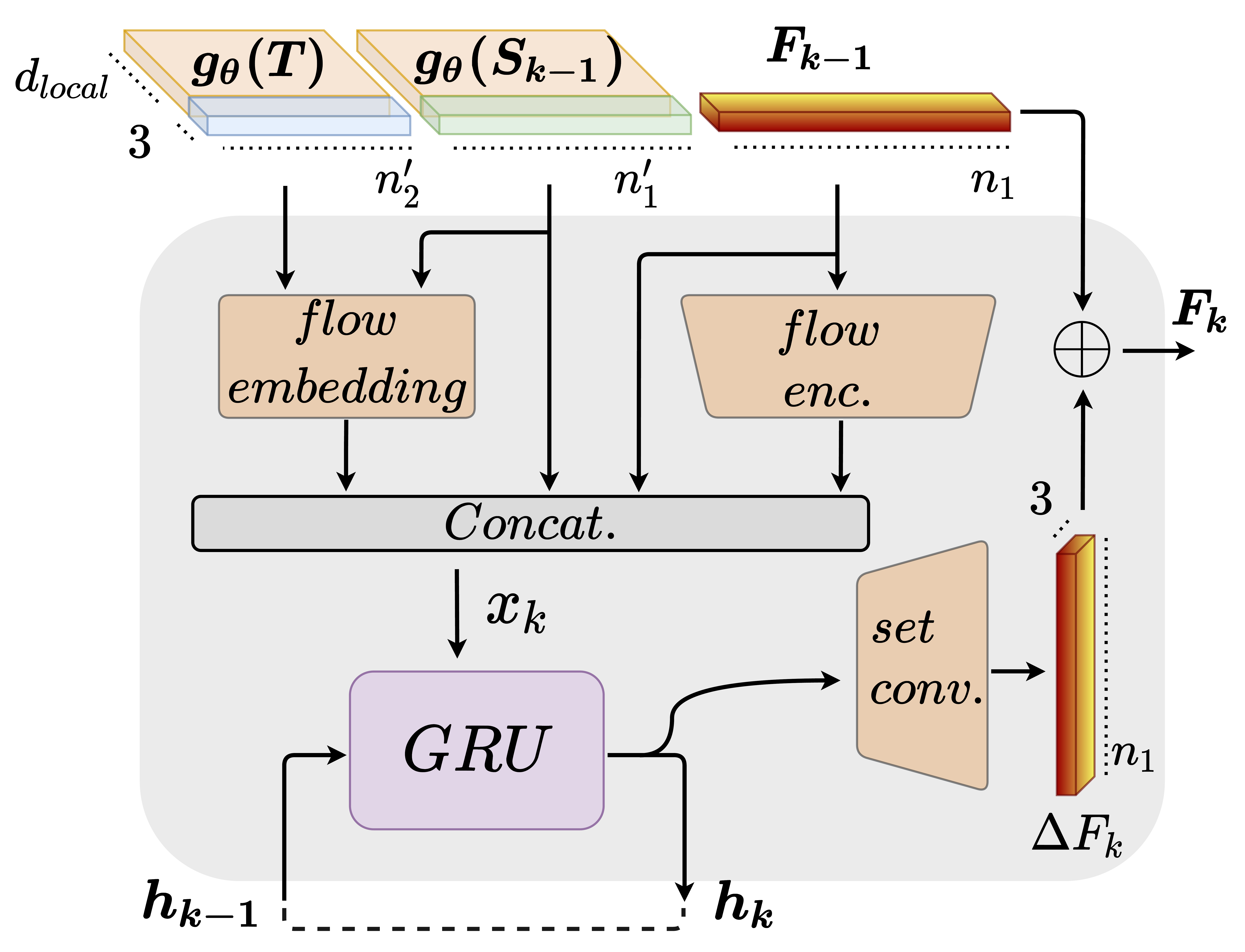}
\end{center}
   \caption{\textbf{Local Update Unit.} The previous iteration predicted flow $F_{k-1}$ and its encoding features, together with local features descriptors of the current state source $g_{\theta}(S_{k-1})$, and of the target $g_{\theta}(T)$, and their correlation ($\textit{flow\_embedding(}\cdot,\cdot\textit{)}$), are concatenated into $x_k$. Gated Recurrent Unit (GRU) followed by \textit{set\_up\_conv} layers produces flow refinement $\Delta F_k$,  as described in Sec. \hyperref[sec:local]{4.3}.}
\label{fig:local_update_unit}
\end{figure}

We define $x_k\in {\mathbb R}^{{n_1'}\times (d_{local} + d_{corr} + 3 + d_{motion})}$ to be the concatenation of the warped source's local features, local flow embedding, previous iteration flow, and previous iteration flow's features.
The previous iteration flow's features are obtained by passing it through two \textit{set\_conv} layers called \textit{flow\_enc}: ${\mathbb R}^{n_1\times 3} \mapsto {\mathbb R}^{n_1'\times d_{motion}}$, as shown in Fig.~\ref{fig:local_update_unit}.

For the initialization of the first iteration's hidden state, we pass the local features of the source point cloud $g_\theta(S)$, through two \textit{set\_conv} layers.

\noindent\textbf{Scene flow prediction.}
\space Given the new hidden state $h_k$ produced by the GRU cell, we use a flow regressor consisting of two \textit{set\_conv} layers to estimate the flow refinement $\Delta F_k$.
The updated flow is then calculated as $\boldsymbol{F_k} = F_{k-1}+\Delta F_k$.

Regression of flow refinements at totally different scales by the same CNN component is challenging. Hence, to encourage our system to learn coarse displacements in first iterations, we multiply the magnitude of each predicted scene flow by a factor $\frac{1}{(C\cdot(k-1) + 1)}$, where C is a hyper-parameter.

\section{Training Loss Functions}
To train our iterative system, we unroll $K$ iterations and apply a loss function for each iteration prediction $\boldsymbol{F}_k$:
\begin{equation}\label{eq:l_seq}
    \mathcal{L}_{seq.}=\sum_{k=1}^{K}\mathcal{L}_k.
\end{equation}
Each iteration loss $\mathcal{L}_k$ in the sequence can be chosen to be a self-supervised (Sec.~\ref{sec:self}) or a fully-supervised (Sec.~\ref{sec:fully}) loss.
\begin{table*}
\centering
\begin{tabular}{@{}l|lccccc@{}}
\toprule
Dataset                         & \multicolumn{1}{l|}{Method}        & \multicolumn{1}{l|}{Sup.}          & \multicolumn{1}{l}{EPE3D$\downarrow$} & \multicolumn{1}{l}{Acc3DS$\uparrow$} & \multicolumn{1}{l}{AccDR$\uparrow$} & \multicolumn{1}{l}{Outliers3D$\downarrow$} \\ \midrule
\multirow{9}{*}{FlyingThings3D} & \multicolumn{1}{l|}{ICP\cite{rigidicp}}           & \multicolumn{1}{c|}{\textit{Self}} & 0.4062                    & 0.1614                     & 0.3038                    & 0.8796                         \\
                                & \multicolumn{1}{l|}{Ego-motion\cite{Ego}}    & \multicolumn{1}{c|}{\textit{Self}} & 0.1696                    & 0.2532                     & 0.5501                    & 0.8046                         \\
                                & \multicolumn{1}{l|}{PointPWC-Net\cite{pointPWC}}  & \multicolumn{1}{c|}{\textit{Self}} & 0.1246                    & 0.3068                     & 0.6552                    & 0.7032                         \\
                                & \multicolumn{1}{l|}{\textbf{Ours}} & \multicolumn{1}{c|}{\textit{\textbf{Self}}} & \textbf{0.0852}            & \textbf{0.5363}            & \textbf{0.8262}           & \textbf{0.4198}                \\ \cmidrule(l){2-7} 
                                & \multicolumn{1}{l|}{FlowNet3D\cite{flownet3d}}     & \multicolumn{1}{c|}{\textit{Full}} & 0.1136                    & 0.4125                     & 0.7706                    & 0.6016                         \\
                                & \multicolumn{1}{l|}{HPLFlowNet\cite{HPLFlowNet}}    & \multicolumn{1}{c|}{\textit{Full}} & 0.0804                    & 0.6144                     & 0.8555                    & 0.4287                         \\
                                & \multicolumn{1}{l|}{PointPWC-Net\cite{pointPWC}}  & \multicolumn{1}{c|}{\textit{Full}} & 0.0588                    & 0.7379                     & 0.9276                    & 0.3424                         \\
                                & \multicolumn{1}{l|}{FLOT\cite{FLOT}}          & \multicolumn{1}{c|}{\textit{Full}} & 0.0520                    & 0.7320                     & 0.9270                    & 0.3570                         \\
                                & \multicolumn{1}{l|}{\textbf{Ours}} & \multicolumn{1}{c|}{\textit{\textbf{Full}}} & \textbf{0.0455}           & \textbf{0.8162}            & \textbf{0.9614}           & \textbf{0.2165}                \\ \midrule
\multirow{9}{*}{KITTI}          & \multicolumn{1}{l|}{ICP\cite{rigidicp}}           & \multicolumn{1}{c|}{\textit{Self}} & 0.5181                    & 0.0669                     & 0.1667                    & 0.8712                         \\
                                & \multicolumn{1}{l|}{Ego-motion\cite{Ego}}    & \multicolumn{1}{c|}{\textit{Self}} & 0.4154                    & 0.2209                     & 0.3721                    & 0.8096                         \\
                                & \multicolumn{1}{l|}{PointPWC-Net\cite{pointPWC}}  & \multicolumn{1}{c|}{\textit{Self}} & 0.2549                    & 0.2379                     & 0.4957                    & 0.6863                         \\
                                & \multicolumn{1}{l|}{\textbf{Ours}} & \multicolumn{1}{c|}{\textit{\textbf{Self}}} & \textbf{0.1021}           & \textbf{0.7080}            & \textbf{0.8394}           & \textbf{0.2456}                 \\ \cmidrule(l){2-7} 
                                & \multicolumn{1}{l|}{FlowNet3D\cite{flownet3d}}     & \multicolumn{1}{c|}{\textit{Full}} & 0.1767                    & 0.3738                     & 0.6677                    & 0.5271                         \\
                                & \multicolumn{1}{l|}{HPLFlowNet\cite{HPLFlowNet}}    & \multicolumn{1}{c|}{\textit{Full}} & 0.1169                    & 0.4783                     & 0.7776                    & 0.4103                         \\
                                & \multicolumn{1}{l|}{PointPWC-Net\cite{pointPWC}}  & \multicolumn{1}{c|}{\textit{Full}} & 0.0694                    & 0.7281                     & 0.8884                    & 0.2648                         \\
                                & \multicolumn{1}{l|}{FLOT\cite{FLOT}}          & \multicolumn{1}{c|}{\textit{Full}} & 0.0560                     & 0.7550                     & 0.9080                    & 0.2420                         \\
                                & \multicolumn{1}{l|}{\textbf{Ours}} & \multicolumn{1}{c|}{\textit{\textbf{Full}}} & \textbf{0.0546}           & \textbf{0.8051}            & \textbf{0.9254}           & \textbf{0.1492}                \\ \bottomrule
\end{tabular}
\vspace{1.8mm} 
   \caption{\textbf{Evaluation results on FlyingThings3D and KITTI datasets.} All methods trained only on FlyingThings3D. \textit{Self /}\textit{ Full} means self-supervised / fully-supervised, where on KITTI \textit{Self /}\textit{ Full} refers to the training type on FlyingThings3D of the respective model that is evaluated on KITTI. Our method outperforms all baselines on all metrics in both fully-supervised and self-supervised training. Our self-supervised version is the only self-supervised method with EPE3D below $10m$ on FlyingThings3D, and it shows its generalization ability by more than $50\%$ improvement over existing baselines on KITTI.}
\label{table:results}
\end{table*}

\subsection{Self-supervised Loss}\label{sec:self}

Due to the lack of labeled data for 3D scene flow, we designed our solution to be trained in a self-supervised manner, i.e. without the need of ground-truth flow.

\noindent\textbf{Chamfer Loss.} 
\space We follow previous works  \cite{3dcoded, pixel2mesh, pointPWC} and choose the Chamfer distance, which enforces the source to move toward the target according to mutual closest points, as our self-supervised data loss:
\begin{multline}
    \mathcal{L}^{ch.}_k=\mathcal{D}_{ch.}(S_k, T)\stackrel{\mathrm{def}}{=} \\
    \sum_{p\in S_k}\min_{q\in T}\|p-q\|_2^2 +\sum_{q\in T}\min_{p\in S_k}\|q-p\|_2^2
\end{multline}
\noindent where $S_k := S + F_k$ is the warped source according to the predicted flow at iteration k.

\noindent\textbf{Regularization Loss.} 
\space Since Chamfer distance has multiple local minima, it is crucial to regularize it in order to reach sufficient convergence. 
Another reason for which our system requires strong regularization is that we warp the source according to the predicted flow before encoding it again. Hence, we need to carefully preserve the objects' structures so that encoding the warped scene will produce meaningful local geometric features (Fig.\ref{fig:reg}).

Motivated by \cite{nicp, 3dcoded, laplacian_mesh, pointPWC}, we propose a strong Laplacian regularization, i.e we enforce the source to preserve its Laplacian when warped according to the predicted flow:
\begin{equation}
\begin{split}
        \mathscr{L}(S+F_k&) \stackrel{\mathrm{\simeq}}{}\mathscr{L}(S) \\
        &\Downarrow\\
         \mathscrbf{L}\boldsymbol{(F}&\boldsymbol{{}_k)\longrightarrow 0.}
\end{split}
\end{equation}

We approximate the Laplacian $ \mathscr{L}(X)$ at a point $x_i \in X$ as its distances from all points $x_j\in \mathcal{N}(x_i)$, where $\mathcal{N}(x_i)$ is a set of points, of size $|\mathcal{N}(x_i)|$, in a neighborhood around $x_i$ defined in a sequel. We use $L_1$ norm for regularization, so that our regularization loss is:
\begin{equation}\label{eq:l_reg}
    \mathcal{L}^{reg.}_k= \frac{1}{n_1}\sum_{p_i\in S}{\frac{1}{|\mathcal{N}(p_i)|}}\sum_{p_j\in \mathcal{N}(p_i)}\|F_k(p_i)-F_k(p_j)\|_1,
\end{equation}
\noindent where $F_k(p_i)$ is the value of the predicted scene flow at point $p_i$, and $n_1$ is the number of points in the source.

To reduce the computational overhead of nearest neighbors for a large $K$, we use $\mathcal{N}(p_i)=\mathcal{N}_a(p_i) \cup \mathcal{N}_b(p_i)$, where $\mathcal{N}_a(p_i)$ is the $K_a$ nearest neighbours of $p_i$, and $\mathcal{N}_b(p_i)$ is calculated by random sampling $K_b$ points in an Euclidean ball around $p_i$, with radius $r_b$.

The overall self-supervised loss is a weighted sum of Chamfer and regularization losses, over all sequence iterations:
\begin{equation}\label{eq:l_self}
    \boldsymbol{\mathcal{L}^{self}}=\sum_{k=1}^{K}\mathcal{L}_k^{self}=\sum_{k=1}^{K}\alpha_k\mathcal{L}^{ch.}_k + \beta_k\mathcal{L}_k^{reg.}
\end{equation}

\subsection{Fully-supervised Loss}\label{sec:fully}
In order to show our architecture efficiency, we further train our system in a fully-supervised manner, using the $L_1$ loss:
\begin{equation}
    \mathcal{L}^{L_1}_k= \frac{1}{n_1}\sum_{p_i\in S}\|F_k(p_i)-F_{GT}(p_i)\|_1,
\end{equation}
\noindent where $F_{GT}(p_i)$ is the value of the ground-truth scene flow at point $p_i$.

Unlike previous methods, we add a laplacian regularization loss to our fully-supervised training to encourage our system to preserve objects' structures and approach toward the target in iterations.
The regularization loss is the same as in the self-supervised case, Eq. \eqref{eq:l_reg}.

The overall fully-supervised loss is a weighted sum of $L_1$ and regularization losses, over all sequence iterations:
\begin{equation}\label{eq:l_fully}
    \boldsymbol{\mathcal{L}^{sv}}=\sum_{k=1}^{K}\mathcal{L}_k^{sv}=\sum_{k=1}^{K}\alpha_k\mathcal{L}^{L_1}_k + \beta_k\mathcal{L}_k^{reg.}.
\end{equation}

\section{Experiments}\label{sec:exp}
Following the experimental setup suggested in \cite{HPLFlowNet, flownet3d, FLOT, Ego, pointPWC}, we first train and evaluate our model on synthetic dataset FlyingThings3D \cite{ft3d} (Sec.~\ref{sec:ft3d}) using both self-supervised and fully-supervised approaches. Then, we test the models' performance on the real-world KITTI scene flow dataset \cite{kitti2015, kitti2018} without any fine-tuning (Sec.~\ref{sec:kitti}).
Finally, in Seq.~\ref{sec:ablation}, we conduct ablation studies regarding the inference iterations number, and the importance of the regularization loss. 

\begin{figure*}
\begin{center}
\includegraphics[width=0.9\linewidth]{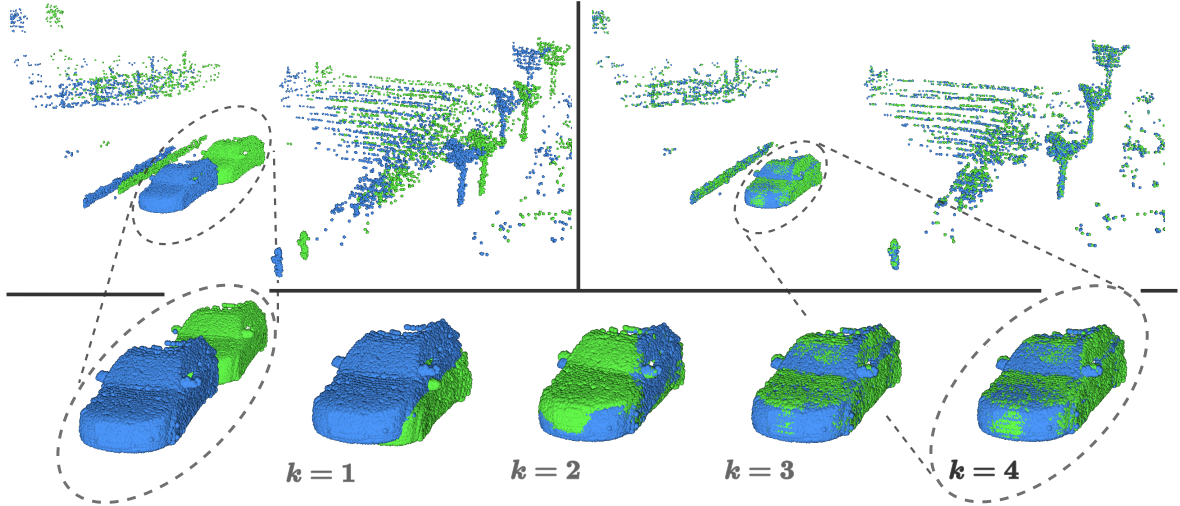}
\end{center}
\vspace{-2.5mm} 
   \caption{\textbf{Inference iterations.} Self-supervised model output example from the KITTI test set. Top-left: input source (green) and target (blue) scans. Top-right:
   overlay of the warped source and the target. Bottom: a closer observation on warped source toward the target during four iterations of flow estimation.}
\label{fig:iters}
\end{figure*}
\begin{figure}[t]
\begin{center}
\includegraphics[width=0.98\linewidth]{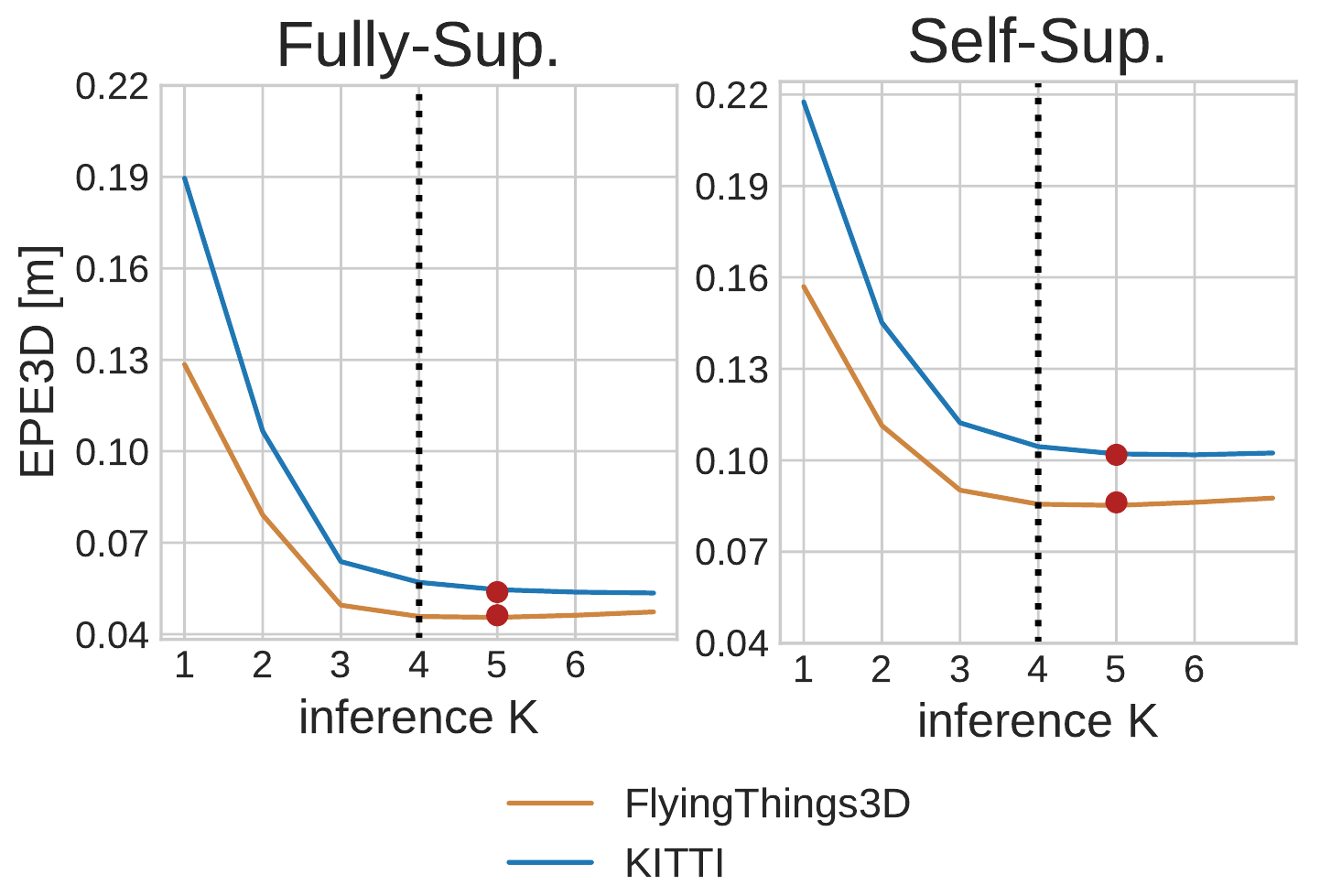}
\end{center}
\vspace{-1.5mm} 
   \caption{\textbf{EPE3D vs. inference K.} For the fully-supervised (left) and the self-supervised (right) trained models, the best $K$ (red) based on the FlyingThings3D validation set (orange), used for test on KITTI (blue). Both models trained with $K=4$ (dashed line).}
\label{fig:graphs}
\end{figure}

\noindent\textbf{Evaluation Metrics.} 
\space We use the same scene flow evaluation metrics proposed by \cite{flownet3d} and adopted by \cite{HPLFlowNet,FLOT, pointPWC}: 
\begin{itemize}
  \item \textbf{EPE3D(m)} average end-point-error $\|F_{pred} - F_{GT}\|_2$ over each point.
  \item \textbf{Acc3DS(0.05)} percentage of points whose \textbf{EPE3D}$<0.05m$ or relative error $<5\%$
  \item \textbf{Acc3DR(0.1)} percentage of points whose \textbf{EPE3D}$<0.1m$ or relative error $<10\%$
  \item \textbf{Outliers3D} percentage of points whose \textbf{EPE3D}$>0.3m$ or relative error $>10\%$
\end{itemize}
\begin{figure*}
\begin{center}
\includegraphics[width=0.95\linewidth]{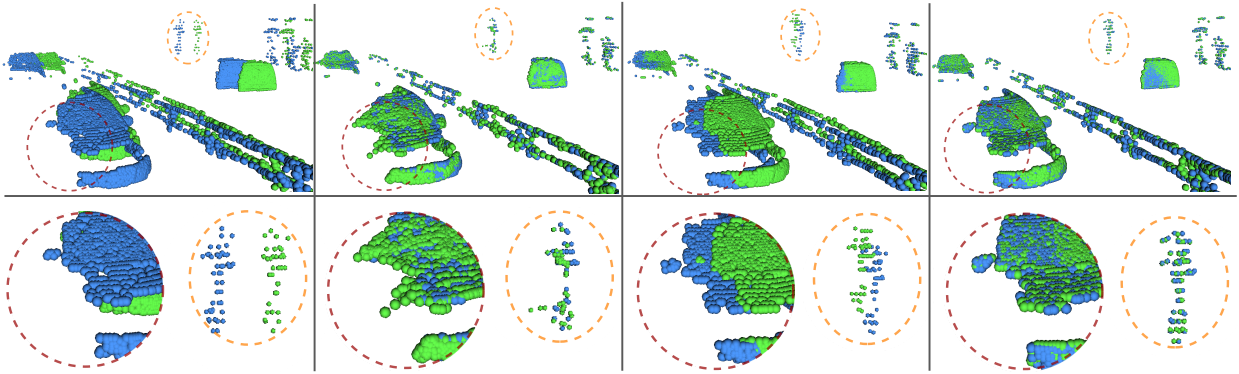}
\end{center}
\vspace{-2mm} 
   \caption{\textbf{Regularization inference example.} An Inference of our self-supervised model pre-trained with three different regularization schemes. From left to right: input source (green) and target (blue) scans, \textit{under}-regularization, \textit{over}-regularization, and
   our \textit{chosen}-regularization. Interesting artifacts are circled and zoomed-in at the bottom. All variations evaluated with $K=4$.} 
\label{fig:reg}
\end{figure*}
\subsection{Evaluation on FlyingThings3D}\label{sec:ft3d}
Due to the difficulty of acquiring dense scene flow data, we follow previous methods \cite{HPLFlowNet, flownet3d, FLOT, pointPWC} and train our system only on the synthetic FlyingThings3D dataset, using the same pre-processing methodology as \cite{HPLFlowNet}.

First, we focus on a self-supervised approach, which does not require any labeled data. Then, to demonstrate our system efficiency, we also conduct experiments using a fully-supervised loss.

\noindent\textbf{Implementation Details.} 
\space The FlyingThings3D dataset contains 19,640 pairs of point clouds in the training set and 3,824 pairs in the validation set.
We first train our system on one quarter of the train data (4910 pairs) and then fine-tune on the full training set, to speed-up training. We used FlyingThings3D validation set for all hyperparameters tuning.

We use $n=8192$ points for each point cloud, batch size of $16$, and unroll $K=4$ iterations at all training procedures, using 8 GTX-2080Ti GPUs.
Pre-training is done for 90 epochs, with a learning rate of $0.002$ and reduced by half at epochs $[50, 70]$.
Self-supervised model is fine-tuned for 30 epochs, with a learning rate of $0.002$ and reduced by half at epochs $[5, 15, 25]$. Fully-supervised model is fine-tuned for 40 epochs, with a learning rate of $0.001$ and reduced by half at epochs $[10, 22, 30]$.

To reduce outliers, we limit the distance of correspondence points to a reasonable displacement range, by zeroing our coarse all-to-all correlation matrix at every entry $(i,j)$ of which $\|p_i-q_j\|_2>10m$.

Lastly, we used FlyingThings3D validation set to determine the best number of iterations for our model at inference time. As discussed in Sec. \ref{sec:ablation}, we set $K=5$ for all tests.

All loss weights $\{\alpha_k\}_{k=1}^K$, $\{\beta_k\}_{k=1}^K$ of all training procedures, and a detailed scheme of our architecture can be found in the supplementary materials.

\noindent\textbf{Results.} 
\space We compare our self-supervised method's results with Iterative-Closest-Point (ICP) \cite{rigidicp}, PointPWC-Net \cite{pointPWC}, and the recent self-supervised method introduced by Tishchenko \etal~\cite{Ego}, and our 
fully-supervised method's results with FlowNet3D \cite{flownet3d}, HPLFlowNet \cite{HPLFlowNet}, PointPWC-Net \cite{pointPWC}, and FLOT \cite{FLOT}.

As shown in Table ~\ref{table:results}, our method outperforms all existing methods on all evaluation metrics on the FlyingThings3D dataset, for both self-supervised and fully-supervised frameworks. Moreover, our self-supervised method is the only self-supervised method with EPE3D below $10m$ on the FlyingThings3D dataset.

\subsection{Generalization on KITTI}\label{sec:kitti}
\space To examine the generalization ability of our method to real-world data, we evaluate a model trained using FlyingThings3D, on real-scans KITTI Scene Flow 2015 \cite{kitti2015, kitti2018} dataset, without any fine-tuning.
Following \cite{HPLFlowNet, pointPWC}, we evaluate our model on all 142 scenes with available 3D data in the training set,
and remove the ground points from the point clouds by height ($<0.3m$).

Our self-supervised method demonstrates a great generalization ability, outperforming all existing self-supervised methods by more than $50\%$.

Our fully-supervised model achieves EPE3D on par with state of the art method \cite{FLOT}, highest accuracy, lowest outliers and benefits from memory efficiency (Fig.~\ref{fig:diagram}).

\begin{table}
\centering
\begin{tabular}{@{}c|cc@{}}
\toprule
Regularization $(\{\alpha_k\},\{\beta_k\})^*$       & EPE3D$\downarrow$ & Outliers3D$\downarrow$ \\ \midrule
\textit{Under}-regularization              & 0.3183            & 0.7698                 \\
\textit{Over}-regularization           & 0.2706            & 0.8941                  \\
\textbf{\textit{Chosen}-regularization} & \textbf{0.1443}   & \textbf{0.3736}        \\ \bottomrule
\end{tabular}
\vspace{1.5mm} 
\caption{\textbf{Regularization.} EPE3D and outliers rate of our self-supervised model pre-trained with different regularization loss weights $\{\alpha_k\},\{\beta_k\}$. All evaluated with $K=4$. It can be seen that both \textit{over}-regularization and \textit{under}-regularization increase errors.
$^*$Exact values of all $\beta_k, \alpha_k$ are in our supplementary.}
\label{table:reg}
\end{table}
\subsection{Ablation Studies}\label{sec:ablation}

\noindent\textbf{Number of iterations.}
\space Although we unrolled four iterations for training, we tested inference with different K values (Fig.~\ref{fig:graphs}). Interestingly, both our models keep slightly improving for a few more iterations on the KITTI test set. On the FlyingThings3D validation set, the models are optimized to one iteration more than the number of training iterations. To decrease run time, one can choose to optimize the system to a smaller number of iterations and get quite good results for 2 or 3 iterations as well. However, a model trained with K=4 not only produces the best results but also benefits from the best generalization ability, especially under large deformations.
Fig.~\ref{fig:iters} shows a qualitative example of our self-supervised method during four inference iterations.

\noindent\textbf{Update unit design choices.}
\space Using a GRU showed better performance than a simple fully-connected layer, increasing the system's generalization ability by $40\%$. 
Regarding its inputs, we found that using the $\textit{flow\_embedding}$ alone as the GRU input increases the validation error by $30\%$.

\noindent\textbf{Regularization.}
\space Since our method re-encodes the warped source at every iteration,
it is crucial to train it using a regularization loss.
While training with under-regularization may distort the objects' structure,
over-regularization may lead to semi-rigid motion predictions, which results in imperfect alignment.
To demonstrate the importance of wisely choosing the regularization loss weights, we pre-train our self-supervised model in three different regularization schemes, changing only the loss weights $\{\beta_k\}_{k=1}^K$, $\{\alpha_k\}_{k=1}^K$, and then evaluate each one of them on the KITTI test set.
We show the quantitative results in Table ~\ref{table:reg}, and a qualitative example in Fig. ~\ref{fig:reg}.
\vspace{-0.4mm}
\section{Conclusions}
In this work, we proposed and studied a novel approach for scene flow estimation by unrolling an iterative scheme using a recurrent architecture that learns the optimal steps toward the solution, called FlowStep3D. We showed the benefit of approaching the solution in a few steps by enforcing strong regularization and re-encoding the warped scene, which is contrary to all previous learning-based solutions.
Experiments performed on synthetic and real LiDAR scans data showed great generalization capability, especially for self-supervised training, improving previous methods by a large margin.

\vspace{3mm}
\noindent\textbf{\small{Acknowledgment.}}
\space \small{This work was partially funded by the Zimin Institute for Engineering Solutions Advancing Better Lives, the Israeli consortiums for soft robotics and autonomous driving, and the Nicholas and Elizabeth Slezak Super Center for Cardiac Research and Biomedical Engineering at Tel Aviv University.}

{\small
\bibliographystyle{ieee_fullname}
\bibliography{egbib}
}

\end{document}